\documentclass[runningheads]{llncs}
\usepackage{graphicx}
\usepackage{amsmath,amssymb,amsfonts}
\usepackage{textcomp}
\usepackage{xcolor}

\usepackage{booktabs}
\usepackage{caption}
\usepackage{subcaption}
\usepackage{graphicx}
\usepackage{multirow}
\usepackage{algorithm}
\usepackage[noend]{algpseudocode}

\usepackage[english]{babel}
\usepackage[utf8]{inputenc}
\usepackage[noend]{algpseudocode}
\DeclareMathOperator*{\argmax}{argmax} % thin space, limits underneath in displays
\begin{document}
\title{Self-supervised Detransformation Autoencoder for Representation Learning in Open Set Recognition}
\author{Jingyun Jia\orcidID{0000-0003-0865-049X} \and
Philip K. Chan\orcidID{0000-0002-3878-4205}}
\authorrunning{J. Jia and P. Chan}
\institute{Florida Institute of Technology, Melbourne FL 32901, USA \\
\email{jiaj2018@my.fit.edu      pkc@fit.edu} }

\maketitle              % typeset the header of the contribution
\begin{abstract}
The objective of Open set recognition (OSR) is to learn a classifier that can reject the unknown samples while classifying the known classes accurately. In this paper, we propose a self-supervision method, Detransformation Autoencoder (DTAE), for the OSR problem. This proposed method engages in learning representations that are invariant to the transformations of the input data. Experiments on several standard image datasets indicate that the pre-training process significantly improves the model performance in the OSR tasks. Moreover, our analysis indicates that DTAE can yield representations that contain some class information even without class labels.

\end{abstract}
\section{Introduction}
\label{sec: intro}
Deep learning has shown great success in recognition and classification tasks in recent years. However, there is still a wide range of challenges when applying deep learning to the real world. Most deep neural networks and other machine learning models are trained under a static close-set scenario. However, the real world is more of an open-set scenario, in which it is difficult to collect samples that exhaust all classes. The problem of rejecting the unknown samples meanwhile accurately classifying the known classes is referred as Open Set Recognition (OSR) \cite{bendale2016towards} or Open Category Learning \cite{DBLP:journals/aim/Dietterich17}. The OSR problem defines a more realistic scenario and has drawn significant attention in applications such as face recognition \cite{DBLP:journals/cviu/OrtizB14}, malware classification \cite{DBLP:journals/corr/abs-1802-04365} and medical diagnoses \cite{DBLP:conf/ipmi/SchleglSWSL17}. 

%Recent advances in self-supervised learning have shown great potential in learning representations without pre-defined annotations. A typical solution is to apply transformations to input images and train a classifier to predict the properties of the transformations instead of the images. 
In this paper, we bring self-supervised pre-training to the OSR problem and fine-tune the pre-trained model with different types of loss functions: classification loss and representation loss. Particularly, we propose      Detransformation Autoencoder (DTAE) for self-supervision. DTAE consists of three components: an encoder, a decoder, and an input transformation module. The encoder encodes all transformed images to representations, and the decoder reconstructs the representations back to the original images before transformations. Compared to the traditional autoencoder, DTAE learns the representations that describe the pixels and are invariant to the transformations. Our contribution in this paper is threefold: First, we introduce DTAE as a self-supervised pre-training method for the OSR tasks. Second, our experiment results show that DTAE significantly improves the model performances for different down-streaming loss functions on several image datasets. Third, our analysis indicates that DTAE is able to capture some cluster information for both known and unknown samples even without class labels.

We organize the paper as follows. In section 2, we give an overview of related work. Section 3 presents the self-supervision method, DTAE, in pre-training for the OSR tasks. Section 4 shows that the pre-training process can significantly improve the model performance in several standard image datasets. Meanwhile, the models pre-trained with DTAE achieve the best performance in detecting the unknown class and classifying the known classes.

\section{Related Work}
\label{sec: relate}
%\subsection{Open set recognition} 
%\cite{DBLP:journals/pami/GengHC21} summarized OSR models to two categories: discriminative models and generative models. Discriminative models can be further divided into traditional machine learning methods and deep neural networks based methods. The generative models consist of instance generation based methods and non-instance generation based methods. %Here, we can divide neural network based OSR techniques into three categories based on the training set compositions. The first category includes the techniques that borrow additional data in the training set. To better discriminate known class and unknown class, \cite{shu2018unseen} and \cite{saito2018open} introduced unlabeled data during the training phase. The research works that generate additional training data fall in the second category of open set recognition techniques. Most data generation methods are based on GANs. \cite{DBLP:conf/bmvc/GeDG17} introduced a conditional GAN to generate some unknown samples followed by OpenMax open set classifier. The third category of open set recognition does not require additional data. Most of the research works require outlier detection for the unknown class. 
 We can divide neural network based OSR techniques into three categories based on the training set compositions. The first category borrows additional data as the unknown samples in the training set. To better discriminate known class and unknown class, Shu et al.\ \cite{shu2018unseen} and Saito et al.\ \cite{saito2018open} introduce unlabeled data during the training phase as the unknown class. The second category generates additional data as the unknown class, Ge et al.\ \cite{DBLP:conf/bmvc/GeDG17} introduce a conditional GAN to generate unknown samples followed by an OpenMax classifier. The third category does not use additional data, Hassen and Chan \cite{DBLP:journals/corr/abs-1802-04365} propose ii loss for the OSR problem. It first finds the representations for the known classes during training and then recognizes an instance as unknown if it does not belong to any known classes. Jia and Chan \cite{DBLP:conf/icann/JiaC21} propose MMF as a loss extension to further separate the known and unknown representations for the OSR problem. CROSR in \cite{DBLP:conf/cvpr/YoshihashiSKYIN19} trains networks for joint classification and reconstruction of the known classes to combine the learned representation and decision in the OSR task. Perera et al.\ \cite{DBLP:conf/cvpr/PereraMJMWOP20} adopt a self-supervision framework to force the network to learn more informative features when separating the unknown class. Specifically, they used the output of the autoencoder as auxiliary features for the OSR task.
 
 Self-supervision in representation learning generally uses a pretext task that is different from the primary task. The pretext task includes reconstructing the input based on a smaller number of features (autoencoders), classifying transformations such as rotations \cite{DBLP:conf/iclr/GidarisSK18}, intra-sample vs inter-sample transformations in contrastive loss \cite{DBLP:conf/icml/ChenK0H20}, redundancy reduction in learned features from transformations \cite{DBLP:conf/icml/ZbontarJMLD21}. In an Autoencoder, the ``labels'' are the input samples themselves, and the network learns the representations of the inputs by minimizing the dissimilarity between the input and output. Denoising autoencoder (DAE) corrupts the input samples first, then the network is trained to denoise corrupted versions of their inputs to reconstruct back to the their original forms. RotNet in \cite{DBLP:conf/iclr/GidarisSK18} uses image rotations to label the dataset, and the model is trained on the classification task of recognizing the rotation classes. SimCLR in \cite{DBLP:conf/icml/ChenK0H20} introduces contrastive loss to improve the quality of learned representations. Given transformed samples, the contrastive loss reduces the intra-sample distances meanwhile increase the inter-sample distances. Barlow twins in \cite{DBLP:conf/icml/ZbontarJMLD21} feeds distorted versions of a sample to two identical networks, and proposes an objective function that makes the cross-correlation matrix between their outputs as close to the identity matrix to minimize the redundancy between components of the representations.
 
 Our proposed method uses a self-supervision approach to learning the features of the known classes without using additional unknown samples. Unlike DAE, our proposed method includes the original input samples in the training process. Moreover, we augment the input samples with different rotation transformations. The network learns the representations that are invariant to the transformations of the input data by decoding all the transformed images back to the original ones before transformations. 

%\subsection{Classification loss functions vs. representation loss functions}

\section{Approach}
\label{sec: app}

\begin{figure}[t]
 \begin{subfigure}[b]{\textwidth}
 \centering
               \includegraphics[width=0.8\linewidth]{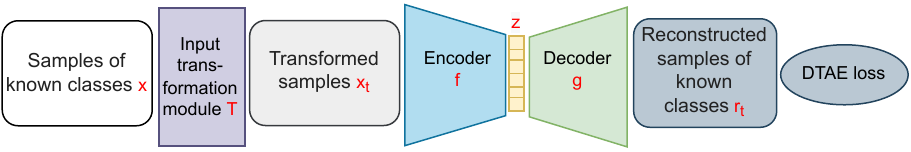}
                \caption{Pre-training with DTAE}
                \label{fig: pre-train}
        \end{subfigure}% 
        \\
  \begin{subfigure}[b]{0.49\textwidth} 
  \centering
  \includegraphics[width=0.95\linewidth]{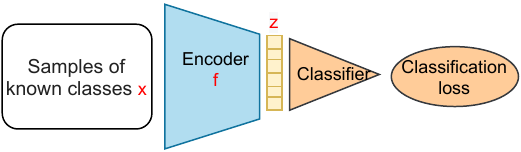}
                \caption{Fine-tuning with classification loss}
                \label{fig: fine-tune-cls}
        \end{subfigure}% 
  \begin{subfigure}[b]{0.49\textwidth} 
  \centering
  \includegraphics[width=0.8\linewidth]{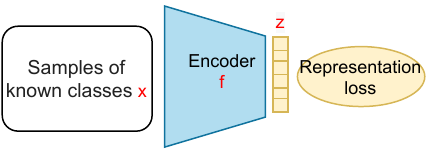}
                \caption{Fine-tuning with representation loss}
                \label{fig: fine-tune-rep}
        \end{subfigure}%  
\caption{The two stage training process of the OSR problem. The first stage is pretraining with DTAE; The second stage is fine-tuning with classification or representation loss.}
\label{fig: de-transformation-autoencoder-training}
\end{figure}

We propose a two-step training process (pre-training step and fine-tuning step) for the OSR problems, thus better separating different classes in the feature space. As illustrated in Figure \ref{fig: de-transformation-autoencoder-training}, the training process includes two steps: 1) pre-training step uses detransformation autoencoder (DTAE) to learn features for all the input data; 2) fine-tuning step uses representation loss functions or classification loss functions to learn discriminative features for different classes. 
% \begin{figure*}[t]

%  \begin{subfigure}{0.45\textwidth}
%               \includegraphics[width=0.98\linewidth]{approach/img/dtae-new.pdf}
%                 \caption{Pre-training}
%                 \label{fig: pre-train}
%         \end{subfigure}% 
%   \begin{subfigure}{0.28\textwidth} 
%   \includegraphics[width=0.9\linewidth]{approach/img/dtae-cls.pdf}
%                 \caption{Fine-tuning with classification loss}
%                 \label{fig: fine-tune-cls}
%         \end{subfigure}% 
%   \begin{subfigure}{0.28\textwidth}                
%   \includegraphics[width=0.9\linewidth]{approach/img/dtae-rep.pdf}
%                 \caption{Fine-tuning with representation loss}
%                 \label{fig: fine-tune-rep}
%         \end{subfigure}% 
    
% \caption{The training process of using detransformation autoencoder.}
% \label{fig: de-transformation-autoencoder-training}
% \end{figure*}

\subsection{Pre-training step}
The objective of the self-supervised pre-training process is to learn some meaningful representations via pretext tasks without semantic annotations. The desirable features should be invariant under input transformations, meanwhile, contain the essential information that can reconstruct the original input. We propose a detransformation autoencoder (DTAE) to learn representations by reconstructing (``detransforming'') the original input from the transformed input. DTAE employs a transformation module and an encoder-decoder structure. While the encoder extracts the representations, the decoder reconstructs the original input from the learned representations. 

The motivation of DTAE is to learn better representations for the OSR problem via encouraging intra-sample similarity and intra-class similarity of the learned representations. For example, if we have samples from ``cat'' class and ``dog'' class, given sample ``cat1'' and its transformation ``cat1a'', we can learn their representations $z_{c1}$ and $z_{c1a}$. Similarly, we can learn the representations of ``cat2'' and ``dog1'' as $z_{c2}$ and $z_{d1}$. The intra-sample similarity describes the similarity between the representations of the original input and its transformations, as $z_{c1}$ and $z_{c1a}$ in our example, and we denote this similarity as $sim(z_{c1}, z_{c1a})$. As the decoder in DTAE reconstructs the \textit{same} original samples from the learned representations of \textit{both} original and transformed samples, the learned representations are of high intra-sample similarity. Thus the learned representations are invariant to the transformations and contain important features of the samples. The intra-class similarity describes the similarity among the learned representations of the same class, as $z_{c1}$ and $z_{c2}$ in our example, and we denote this similarity as $sim(z_{c1}, z_{c2})$. The encoder-decoder structure in DTAE is a generative model that embeds crucial features in lower dimensions. Compared to a discriminate model, the representations learned by a generative model contain more comprehensive information to reconstruct the inputs. Thus, for a generative model, the learned representations of samples from the same class should be more similar than those of different classes. Overall, the desired representation space should satisfy $sim(z_{c1}, z_{c1a}) > sim(z_{c1}, z_{c2}) > sim(z_{c1}, z_{d1})$.

As shown in Figure \ref{fig: pre-train}, in the pre-training stage with DTAE, the input transformation module $T$ transforms any given data example $x$ to several correlated views of the same example, denoted as $x_t = T(x)$.  The network-based encoder $f(\cdot)$ extracts representation vectors from transformed data examples. Furthermore, decoder $g(\cdot)$ reconstructs the original data examples from the representation vectors. Let $r_t$ denotes the reconstructed data example from transformed input $x_t$, then the detransformation loss function becomes:

\begin{equation}
    \mathcal{L} (x, r_t) = \mathcal{L} (x, g(f(x_t)))
\end{equation}
where $r_t = g(f(x_t))$. Specifically, we use MSE (Mean Squared Error) loss and have a total of $M$ transformations, the loss function can be defined as:

\begin{equation}
    \mathcal{L}_\text{DTAE} = \frac{1}{2} \sum^{M-1}_{t=0}\sum^N_{i=1}(x_i - r_{it})^2 
\label{eq: detrans}
\end{equation}
Each of the $N$ data points has $M$ transformations, and there are $M \times N$ data points after the input transformation module. In this work, we consider four transformations for each data example, i.e. $t \in \{0, 1, 2, 3\}$ for all $N$ input examples, resulting in $4N$ data points. For this paper, the four transformations in our experiments are rotations of an image: 0, 90, 180, and 270 degrees.

\subsection{Fine-tuning step}

While the pre-trained network can be fine-tuned by different loss functions, we focus on two types of loss functions in this paper: the classification loss and the representation loss. The objective of classification loss is to lower the classification error of the training data explicitly in the decision layers. One of the widely used classification loss functions is cross-entropy loss. The objective of representation loss functions is to learn better representations of training data. The representation loss functions are normally applied to the representation layers, such as triplet loss \cite{DBLP:conf/cvpr/SchroffKP15} and ii loss \cite{DBLP:journals/corr/abs-1802-04365}. %Triplet loss intends to find an embedding space where the distance between an anchor instance and another instance from the same class is smaller by a user-specified margin than the distance between the anchor instance and another instance from a different class. The objective of ii loss is to maximize the distance between different classes (inter-class separation) and minimize the distance of an instance from its class mean (intra-class spread). So that in the learned representation, instances from the same class are close to each other while those from different classes are further apart.

The fine-tuning network shares the same encoder and representation layer with the pre-training network. However, compared with the pre-training process, the fine-tuning process does not contain the input transformation module, which means the training examples are sent directly into the encoder. Moreover, instead of connecting to a decoder, the representation layer connects to a classification loss function or a representation loss function as shown in Figure \ref{fig: fine-tune-cls} and Figure \ref{fig: fine-tune-rep}. In this work, we consider both classification loss (cross-entropy loss) and representation loss (triplet loss \cite{DBLP:conf/cvpr/SchroffKP15} and ii loss \cite{DBLP:journals/corr/abs-1802-04365}) in the OSR task. 

\subsection{Open Set Recognition (OSR)}
\label{sec:osr}
A typical OSR task solves two problems: classifying the known classes and identifying the unknown class. From the representation level, the instances from the same class are close to each other, while those from different classes are further apart. Under this property, we propose the outlier score:

\begin{equation}\label{eq:outlier}
outlier\_score(x) = \min_{1\leq{i}\leq{C}}\|\mu_i-z\|_{2}^2,
\end{equation}
Where $z$ is the learned representation of test sample $x$, $\mu_i$ is the representation centroid of the known class $i$. There are multiple ways to set the outlier threshold. Here, we sort the outlier score of the training date in ascending order and pick the 99 percentile outlier score value as the outlier threshold. Then, for the $C$ known classes, we predict the class probability $P(y=i | x)$ for each class. When a network is trained on classification loss, the $P(y=i| x)$ is the output of the classification layer. Whereas in the case of a network without classification layer such as Figure \ref{fig: fine-tune-rep}, we calculate $P(y=i | x)$ as:

\begin{align}\label{eq: prob}
P(y=i | x) = \frac{e^{-\|\mu_i-z\|_{2}^2}}{\sum_{j=1}^{C}e^{-\|\mu_j-z\|_{2}^2}}
\end{align}

In summary, a test instance is recognized as ``unknown'' if its outlier score is greater than the threshold $t$, otherwise it is classified as the known class with the highest class probability:

\begin{equation}\label{eq:intra}
    y=
    \begin{cases}
      unknown,& \text{if$\ outlier\_score(x) > t$} \\
      \argmax\limits_{1\leq{i}\leq{C}} P(y=i | x),& \text{otherwise}
    \end{cases}
\end{equation}

% \begin{table}
% \caption{The comparison of the training time (in seconds) for the self-supervision methods in different datasets.}
%     \centering
% \scalebox{1.0}{
% \begin{tabular}{l ccc}
% \toprule
%  & AE & RotNet & DTAE \\ \midrule
% MNIST & 75 & 132 & 137 \\
% Fashion-MNIST & 70 & 118 & 145 \\
% CIFAR-10 & 86 & 147 & 182  \\ \bottomrule
% \end{tabular}}

%     \label{tab: time}
% \end{table}
\section{Experimental Evaluation}
\label{sec: exp}

We evaluate the proposed pre-training method: Detransformation Autoencoder (DTAE) with simulated open-set datasets from the following datasets.

\noindent\textbf{MNIST}  \cite{DBLP:journals/corr/abs-1802-10135} contains 60,000 training and 10,000 testing handwritten digits from 0 to 9,  which is 10 classes in total. Each example is a 28x28 grayscale image. To simulate an open-set dataset, we randomly pick six digits as the known classes participant in the training, while the rest are treated as the unknown class only existing in the test set.

\noindent\textbf{Fashion-MNIST} \cite{xiao2017/online} is associated with 10 classes of clothing images. It contains 60,000 training and 10,000 testing examples. Same as the MNIST dataset, each example is a 28x28 grayscale image. To simulate an open-set dataset, we randomly pick six digits as the known classes participant in the training, while the rest are treated as the unknown class for testing.

\noindent\textbf{CIFAR-10}  \cite{krizhevsky2009learning} contains 60,000 32x32 color images in 10 classes, with 6,000 images per class. There are 50,000 training images and 10,000 test images. We first convert the color images to grayscale and randomly pick six classes out of the ten classes as the known classes, while the remaining classes are treated as the known class only existing in the test set. 

%\noindent\textbf{STL-10} \citep{DBLP:journals/jmlr/CoatesNL11} is an image recognition dataset contains both labeled and unlabeled color images. We use the labeled ones for our experiments. The labeled images include ten classes, with 500 images per class. All the images are of size 96x96. Like the CIFAR-10 dataset, we first convert the color images to grayscale, randomly pick six classes out of the ten classes as the known classes, while the rest are considered unknowns only used in the test set.

\subsection{Evaluation Network Architectures and Evaluation Criteria}
In the proposed method, we use self-supervision in the pre-training stage, and then in the second stage, we fine-tune the pre-trained model with two types of loss functions: classification loss and representation loss. Specifically, we use the cross-entropy loss as the example of classification loss, and use ii loss \cite{DBLP:journals/corr/abs-1802-04365} and triplet loss \cite{DBLP:conf/cvpr/SchroffKP15} as the examples of representation loss. We first trained the model from scratch as a baseline (no pre-training) for each loss function and compared it with the corresponding fine-tuned models after self-supervision. Second, to evaluate our proposed self-supervision technique DTAE, we compare the model performance using DTAE with traditional Autoencoder (AE) and RotNet \cite{DBLP:conf/iclr/GidarisSK18} in the pre-training stage. We also compare the proposed method with OpenMax \cite{bendale2016towards} to show that it is effective to OSR problems.

Figure \ref{fig: pre-train} illustrates the network architecture of the DTAE. Moreover, the hyper-parameters are different based on datasets. For the encoder of the MNIST and the Fashion-MNIST datasets, the padded input layer is of size (32, 32), followed by two non-linear convolutional layers with 32 and 64 nodes. We also use the max-polling layers with kernel size (3, 3) and strides (2, 2) after each convolutional layer. We use two fully connected non-linear layers with 256 and 128 hidden units after the convolutional component. Furthermore, the representation layer is six dimensions in our experiments. The representation layer is followed by a decoder, which is the reverse of the encoder in our experiments. We use the Relu activation function and set the Dropout's keep probability as 0.2. We use Adam optimizer with a learning rate of 0.001. The encoder network architecture of the CIFAR-10 experiment is similar to the MNIST dataset, except the padded input layer is of size (36, 36). We use batch normalization in all the layers to prevent features from getting excessively large. And as mentioned in section \ref{sec:osr}, we use contamination ratio of 0.01 for the threshold selection. The encoder and representation layer maintain the same architecture and hyper-parameters in the fine-tuning network. Meanwhile, the decoder is replaced with different fully connected layers associated with different loss functions. 

We simulate three different groups of open sets for each dataset then repeat each group 10 runs, so each dataset has 30 runs in total. When measuring the model performance, we use the average AUC scores under 10\% and 100\% FPR (False Positive Rate) for recognizing the unknown class. We chose the 10\% FPR limit as higher FPR is generally undesirable, particularly when negative instances are much more abundant than positive instances. We measure the F1 scores for known and unknown classes separately as one of the OSR tasks is to classify the known classes. Moreover, we perform t-tests with 95\% confidence in the AUC scores and F1 scores to see if the proposed DTAE pre-training method can significantly improve different loss functions.

\begin{table*}[t]
%\begin{adjustbox}{width=0.85\columnwidth,center}
\caption{The average ROC AUC scores of 30 runs at 100\% and 10\% FPR of OpenMax and a group of 5 methods (without pre-training as baseline, pre-training with AE, RotNet, DTAE and TAE) for each of the 3 loss functions (ce, ii, triplet). The underlined values are statistically significantly better than the baselines via t-test with 95\% confidence. The values in bold and in brackets are the highest and the second-highest values in each group.}
\centering
\resizebox{0.7\textwidth}{!}{%
\begin{tabular}{l l cc cc cc cc}
\toprule
 & \multicolumn{1}{c}{}  & \multicolumn{2}{c}{MNIST}  & \multicolumn{2}{c}{Fashion-MNIST} & \multicolumn{2}{c}{CIFAR-10} \\ 
 FPR &  & 100\% & 10\%  & 100\% & 10\% & 100\% & 10\% \\ \cmidrule(l){3-4} \cmidrule(l){5-6} \cmidrule(l){7-8}  
 \multirow{1}{*}{OpenMax} &  & 0.9138 & 0.0590 & 0.7405  & 0.0160 & 0.6750 & 0.0060   \\ \midrule
\multirow{5}{*}{ce}   & No pre-training & 0.9255 & 0.0765 & 0.7175 & 0.0300 & 0.5803 & 0.0070  \\ & AE & 0.9410 & \textbf{0.0805} & 0.7346 & 0.0300 & \underline{0.6114} & [0.0084]  \\
& RotNet & 0.9367 & 0.0769 & 0.7364 & [0.0316] & [0.6124] & \underline{0.0083} \\
& DTAE (ours) & \textbf{\underline{0.9523}} & [0.0801] & \textbf{\underline{0.7490}} & \textbf{0.0324} & \textbf{\underline{0.6183}} & \textbf{\underline{0.0086}} \\
& TAE & [\underline{0.9477}] & 0.0799 & [\underline{0.7389}] & 0.0298 & \underline{0.6012} & 0.0075 \\
\midrule
\multirow{5}{*}{ii} & No pre-training & \textbf{0.9578} & 0.0821 & 0.7684 & 0.0399 & 0.6392 & 0.0103 \\  & AE & 0.9560 & \textbf{0.0828} & 0.7636 & 0.0377 & 0.6320 & 0.0098 \\
& RotNet & 0.9530 & 0.0813 & [0.7703] & [0.0404] & [0.6478] & [0.0106] \\ 
& DTAE (ours) & [0.9566] & [0.0825] & \textbf{0.7802 } & \textbf{0.0410} & \textbf{\underline{0.6520}} & \textbf{0.0108} \\ & TAE & 0.9515 & 0.0815 & 0.7657 & 0.0387 & 0.6214 & 0.0091 \\ \midrule
\multirow{5}{*}{triplet} & No pre-training & 0.9496 & 0.0750 & 0.7160 & 0.0211 & 0.6106 & 0.0089 \\ 
& AE & \textbf{0.9563} & \textbf{0.0772} & 0.7254 & 0.0220 & 0.6251 & 0.0090 \\
& RotNet & 0.9342 & 0.0702 & [\underline{0.7435}] & \textbf{\underline{0.0252}} & [\underline{0.6285}] & [0.0095] \\ 
& DTAE (ours) & [0.9543] & [\underline{0.0758}] & \textbf{\underline{0.7441}} & [\underline{0.0234}] & \textbf{\underline{0.6327}} & \textbf{0.0096} \\
& TAE & 0.9531 & 0.0757 & 0.7271 & 0.0215 & 0.6114 & 0.0081 \\
\bottomrule
\end{tabular}}
\label{tab:auc}
%\end{adjustbox}
\end{table*}

\subsection{Experimental Results}

\noindent\textbf{Model performance} We compare the model performances of cross-entropy loss, ii loss, and triplet loss with and without pre-training. Table \ref{tab:auc} are the averaged ROC AUC scores of the model performances in three datasets under different FPR values. Comparing ``RotNet'', ``DTAE'' and ``AE'' rows with ``No pre-training'' rows, we observe that using self-supervision techniques for pre-training significantly improved the model performance. The results also show that our proposed self-supervision method DTAE achieves the top two ROC AUC scores for all the cases. Moreover, with our proposed pre-training method, all three loss functions perform better than OpenMax in 5 out of 6 cases (3 datasets$ \times $2 FPR limits). 

To evaluate the detransformation component of DTAE, we performed an ablation study on our method without detransformation, which is denoted as TAE. Although both DTAE and TAE use transformed instances as input, TAE reconstructs the transformed instances as output, while DTAE reconstructs the original instances as output. Comparing the ``TAE'' rows and ``DTAE'' rows, we observe that the detransformation component in DTAE plays a key role in improving the model performance. That is, our results indicate that learning features invariant to transformations, via detransformation, can yield more effective features than those learned from reconstructing the same samples.

\begin{table*}[t]
%\begin{adjustbox}{width=0.85\columnwidth,center}
\caption{The average F1 scores of 30 runs of OpenMax and a group of 5 methods (without pre-training as baseline, pre-training with AE, RotNet, DTAE and TAE) for each of the 3 loss functions (ce, ii, triplet). The underlined values show statistically significant improvements (t-test with 95\% confidence) comparing to the baselines. The values in bold and in brackets are the highest and the second highest values in each group.}
\centering
\resizebox{0.85\textwidth}{!}{%
\begin{tabular}{l l ccc ccc ccc ccc}
\toprule
 & \multicolumn{1}{c}{}  & \multicolumn{3}{c}{MNIST}  & \multicolumn{3}{c}{Fashion-MNIST} & \multicolumn{3}{c}{CIFAR-10} \\ \midrule
 & & Known & Unknown & Overall & Known & Unknown & Overall & Known & Unknown & Overall\\ \cmidrule(l){3-5} \cmidrule(l){6-8} \cmidrule(l){9-11}  
 \multirow{1}{*}{OpenMax} &  & 0.8964 & 0.7910 & 0.8814 & 0.7473 & 0.5211 & 0.7150 & 0.6456 & 0.5407 & 0.6307  \\ \midrule
\multirow{5}{*}{ce}   & No pre-training & 0.7596 & 0.7561 & 0.7591 & 0.6858 & 0.5591 & 0.6677 & 0.5672 & 0.3697 & 0.5390  \\ & AE & 0.7735 & 0.7894 & 0.7757 & 0.7264 & 0.5481 & 0.7009 & 0.5729 & \underline{0.4605} & [0.5569]   \\
& RotNet & [\underline{0.8931}] & [\underline{0.8447}] & [\underline{0.8862}] & 0.7117 & \textbf{0.5694} & 0.6914 & 0.5616 & \textbf{\underline{0.4729}} & 0.5489  \\
& DTAE (ours) & \textbf{\underline{0.8967}} & \textbf{\underline{0.8579}} & \textbf{\underline{0.8912}} & [\underline{0.7335}] & [0.5692] & [\underline{0.7100}] & \textbf{\underline{0.5911}} & [\underline{0.4728}] & \textbf{\underline{0.5742}} \\
& TAE & \underline{0.8804} & \underline{0.8420} & \underline{0.8749} & \textbf{\underline{0.7482}} & 0.5364 & \textbf{\underline{0.7179}} & [0.5815] & 0.3889 & 0.5540  \\
\midrule
\multirow{5}{*}{ii} & No pre-training & 0.9320 & 0.8833 & 0.9250 & 0.7720 & 0.5870 & 0.7456 & 0.6206 & 0.3570 & 0.5829 \\  & AE & \textbf{0.9387} & \textbf{0.8950} & \textbf{0.9325} & 0.7669 & 0.5745 & 0.7394 & 0.6241 & 0.2527 & 0.5711 
\\
& RotNet & 0.9300 & 0.8761 & 0.9223 & \textbf{0.7771} & \textbf{0.6108} & \textbf{0.7533} & \textbf{\underline{0.6442}} & [\underline{0.3980}] & [\underline{0.6090}]  \\ 
& DTAE (ours) & [0.9344] & [0.8885] & [0.9279] & [0.7768] & [0.6064] & [0.7524] & [\underline{0.6421}] & \textbf{\underline{0.4252}} & \textbf{\underline{0.6111}}  \\ & TAE & 0.9308 & 0.8830 & 0.9240 & 0.7625 & 0.5869 & 0.7374 & 0.6135 & 0.2103 & 0.5559  \\ \midrule
\multirow{5}{*}{triplet} & No pre-training & 0.9103 & 0.8302 & 0.8989 & 0.7491 & 0.5055 & [0.7208] & 0.5798 & 0.4515 & 0.5614 \\ 
& AE & [0.9144] & 0.8356 & [0.9032] & 0.7505 & 0.5051 & 0.7154 & \underline{[0.6086]} & [0.4800] & \underline{0.5902}  
\\
& RotNet & 0.9012 & 0.8182 & 0.8893 & [0.7514] & [0.5376] & [0.7208] & \underline{0.6037} & \textbf{0.4978} & [\underline{0.5886}] \\ 
& DTAE (ours) & \textbf{0.9166} & \textbf{0.8513} & \textbf{0.9073} & \textbf{0.7558} & \textbf{\underline{0.5459}} & \textbf{\underline{0.7259}} & \textbf{\underline{0.6205}} & 0.4724 & \textbf{\underline{0.5993}} 
\\& TAE & 0.9126 & [0.8387] & 0.9021 & 0.7472 & 0.5092 & 0.7132 & 0.5926 & 0.4220 & 0.5682  \\
\bottomrule
\end{tabular}}
\label{tab:f1}
%\end{adjustbox}
\end{table*}

Table \ref{tab:f1} shows that the OSR performances of different methods are measured by F1 scores in known and unknown class domains. We first calculate the F1 scores for each known class and the unknown class, then average all the classes as the Overall F1 scores. The results show that models with pre-training achieve statistically significant improvements. Moreover, Our proposed method also achieves the top two F1 scores in 26 out of 27 cases (3 loss functions$\times$3 datasets$\times$3 domains).

\begin{figure}[t]

    \begin{minipage}{0.45\textwidth}
\centering
\captionof{table}{The training time (in seconds) for the self-supervision methods in different datasets.}
\scalebox{0.75}{
\begin{tabular}{l ccc}
\toprule
 & AE & RotNet & DTAE \\ \midrule
MNIST & 75 & 132 & 137 \\
Fashion-MNIST & 70 & 118 & 145 \\
CIFAR-10 & 86 & 147 & 182  \\ \bottomrule
\end{tabular}
}
\label{tab: time}
\end{minipage}
\hfill
  \begin{minipage}{0.4\textwidth}
 \centering
            \includegraphics[width=\linewidth,height=3cm]{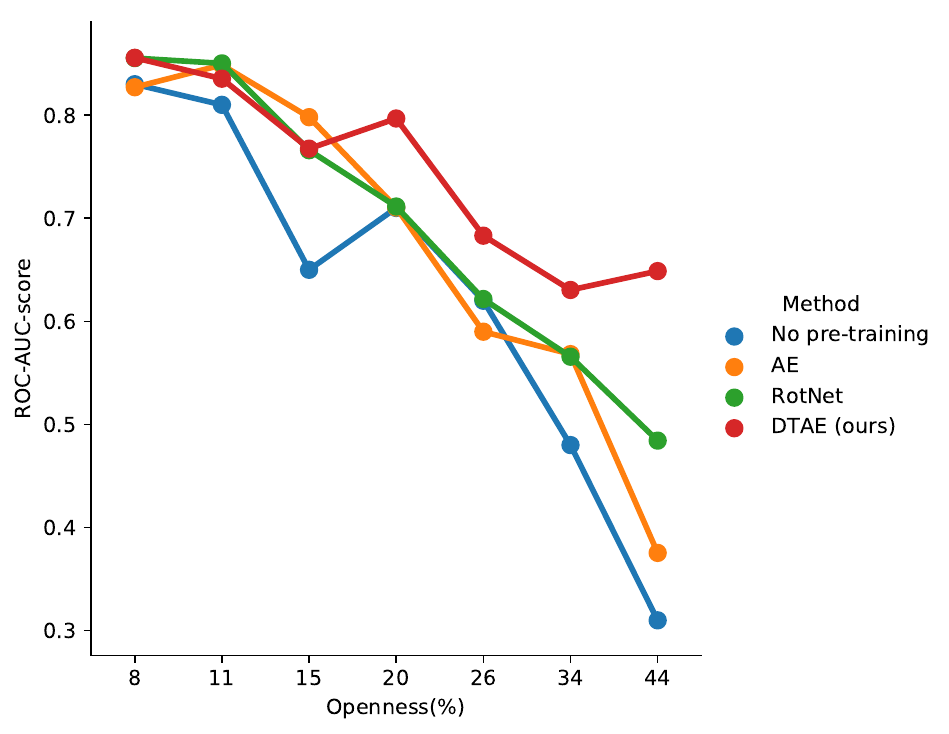}
\caption{AUC-ROC scores against varying Openness.}
\label{fig: openness}
  \end{minipage}
\end{figure}

\noindent\textbf{Training time} While the pre-training step benefits the model performances and does not affect the final model complexity and inference time, it takes extra time during the training phase. Table \ref{tab: time} shows the comparison of the training time of the self-supervised networks in different datasets via NVIDIA RTX 2080. Because RotNet and DTAE both include transformed data as input, they took a longer training time than AE. We observe that DTAE takes a slightly longer training time than RotNet. The reason is that the network structure of DTAE is more complex than that of RotNet. While both RotNet and DTAE share the same encoder and representation layer structures, RotNet uses a softmax layer after the representation layer. Meanwhile, DTAE connects the representation layer with a decoder module. The decoder is the reverse of the encoder, which contains more layers than a softmax layer and needs a longer time in the forward and backward propagations.

\noindent\textbf{Openness study} We also study the model performances against vary Openness \cite{6365193}. Let $n_{train}$ be the number of known classes participant in the training phase, with $n_{test}$ denotes the number of classes in the test set, and $n_{target}$ denotes the number of classes to be recognized in the testing phase. Openness can be defined as follows:

\begin{equation}
    Openness = 1 - \sqrt{\frac{2 \times n_{train}}{n_{test} + n_{target}}}
\end{equation}

 In our experiments with the Fashion-MNIST dataset, we use all the ten classes in testing phase ($n_{test} = 10$) and varying the number of known classes from 2 to 9 ($n_{train} = 2,\dots,9$) in the training phase, and remaining classes together are treated as the unknown class to be recognized along with the known classes during inference ($n_{target} = n_{train} + 1$). That is, the openness is varied from 8\% to 44\%. We evaluate the AUC ROC scores of different models using cross-entropy loss: without pre-training (baseline), pre-training with AE, pre-training with RotNet, and pre-training with our proposed DTAE. The results are shown in Figure \ref{fig: openness}. We observe that the three different models have similar performances when the openness is small. However, the AUC ROC scores of the baseline (No pre-training) degrade rapidly as the Openness increases. Moreover, the trend is alleviated by pre-training with self-supervision methods. Overall, the model pre-trained with DTAE is relatively more robust to openness and achieves the best performance.

\subsection{Analysis}

\begin{figure*}[t]
 \begin{subfigure}[b]{\textwidth}
 \centering
               \includegraphics[width=0.49\linewidth]{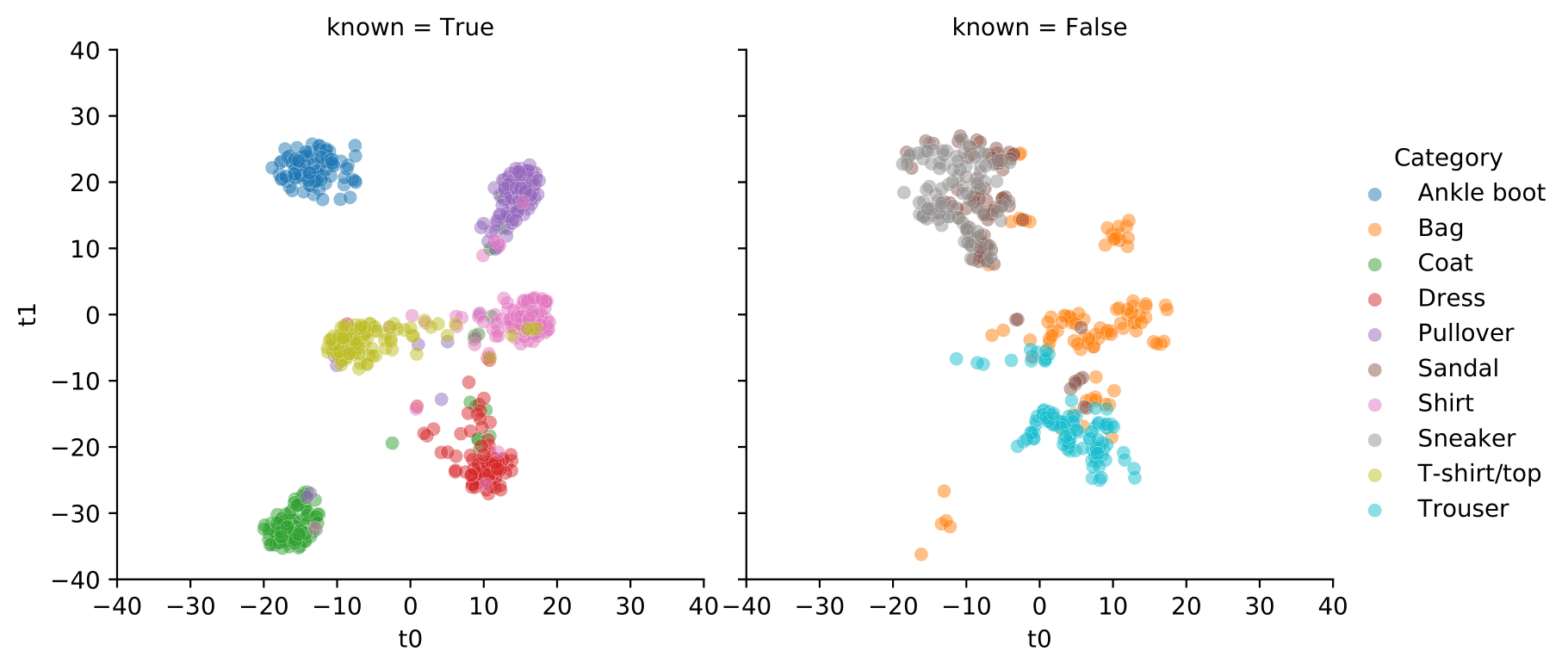}
                \caption{Without pre-training (CE)}
                \label{fig: fmnist-wo-pt-tsne}
        \end{subfigure} \\
 \begin{subfigure}[b]{0.49\textwidth}
               \includegraphics[width=\linewidth]{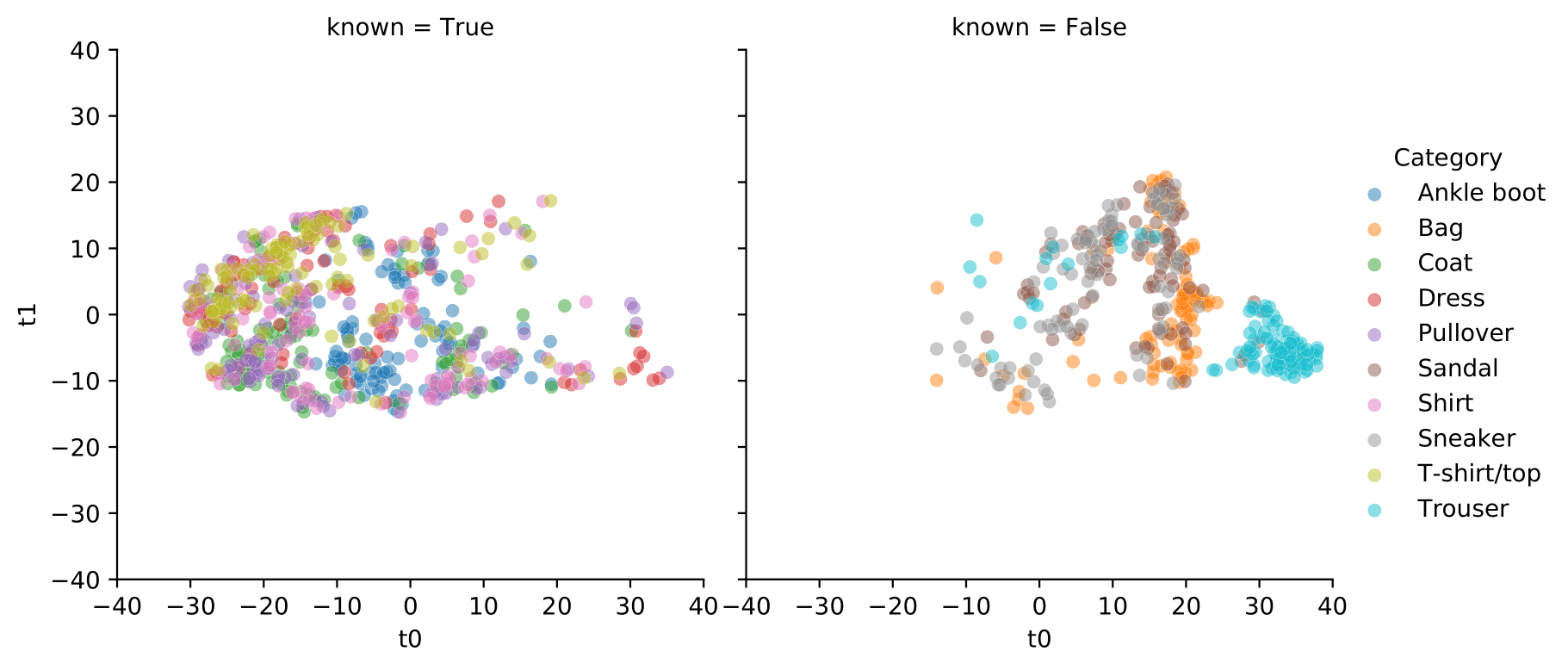}
                \caption{After pre-training (RotNet)}
                \label{fig: fmnist-pt-rotnet-tsne}
        \end{subfigure} 
  \begin{subfigure}[b]{0.49\textwidth}                \includegraphics[width=\linewidth]{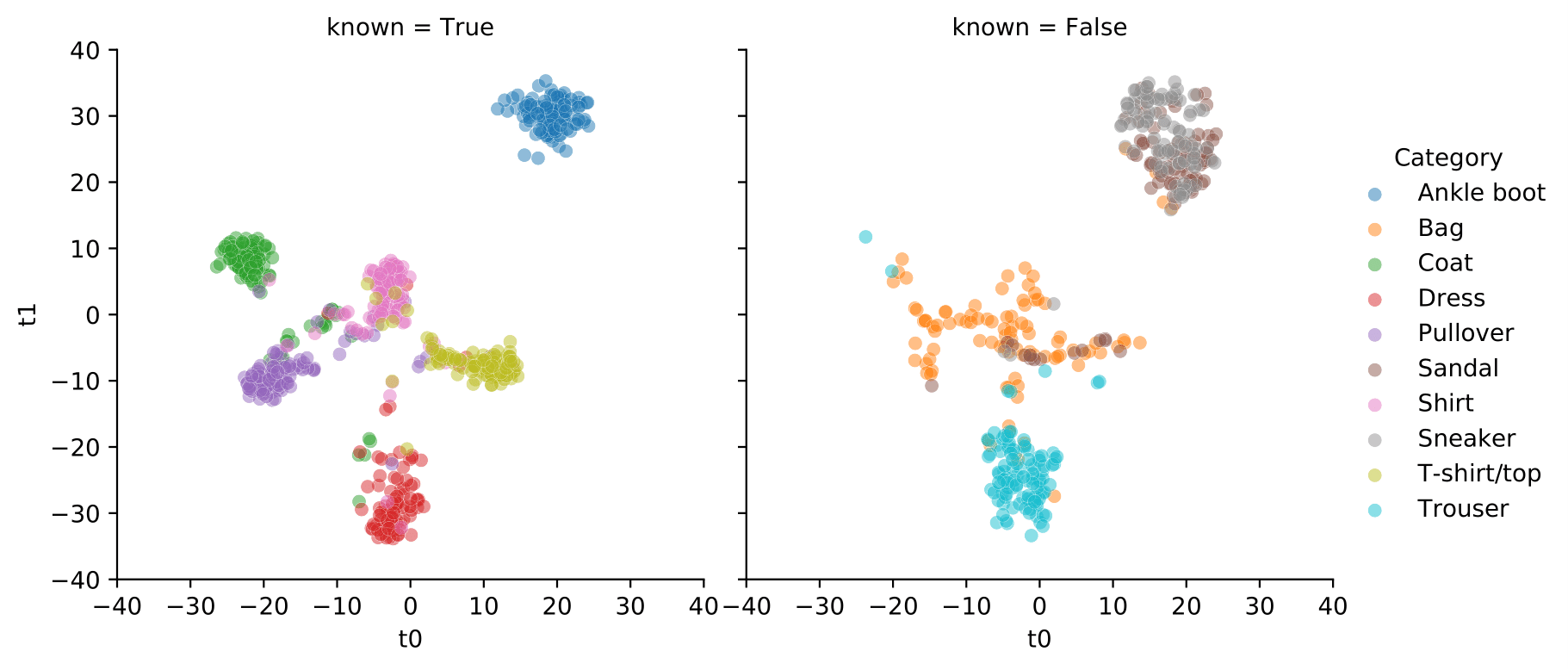}
                \caption{After fine-tuning (RotNet + CE)}
                \label{fig: fmnist-ft-rotnet-tsne}
        \end{subfigure}% 
 \\ \begin{subfigure}[b]{0.49\textwidth}
                \includegraphics[width=\linewidth]{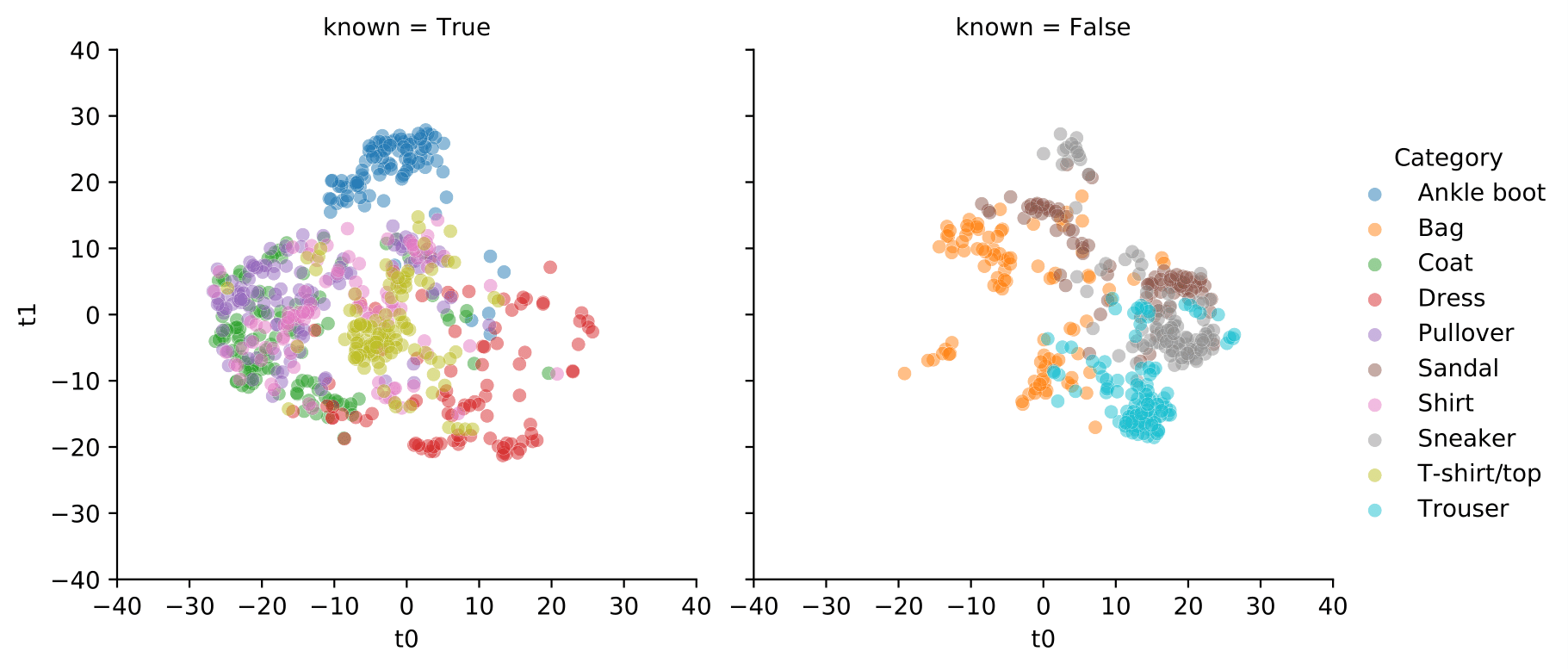}
                \caption{After pre-training (DTAE)}
                \label{fig: fmnist-pt-dtae-tsne}
        \end{subfigure}%
  \begin{subfigure}[b]{0.49\textwidth}
                \includegraphics[width=\linewidth]{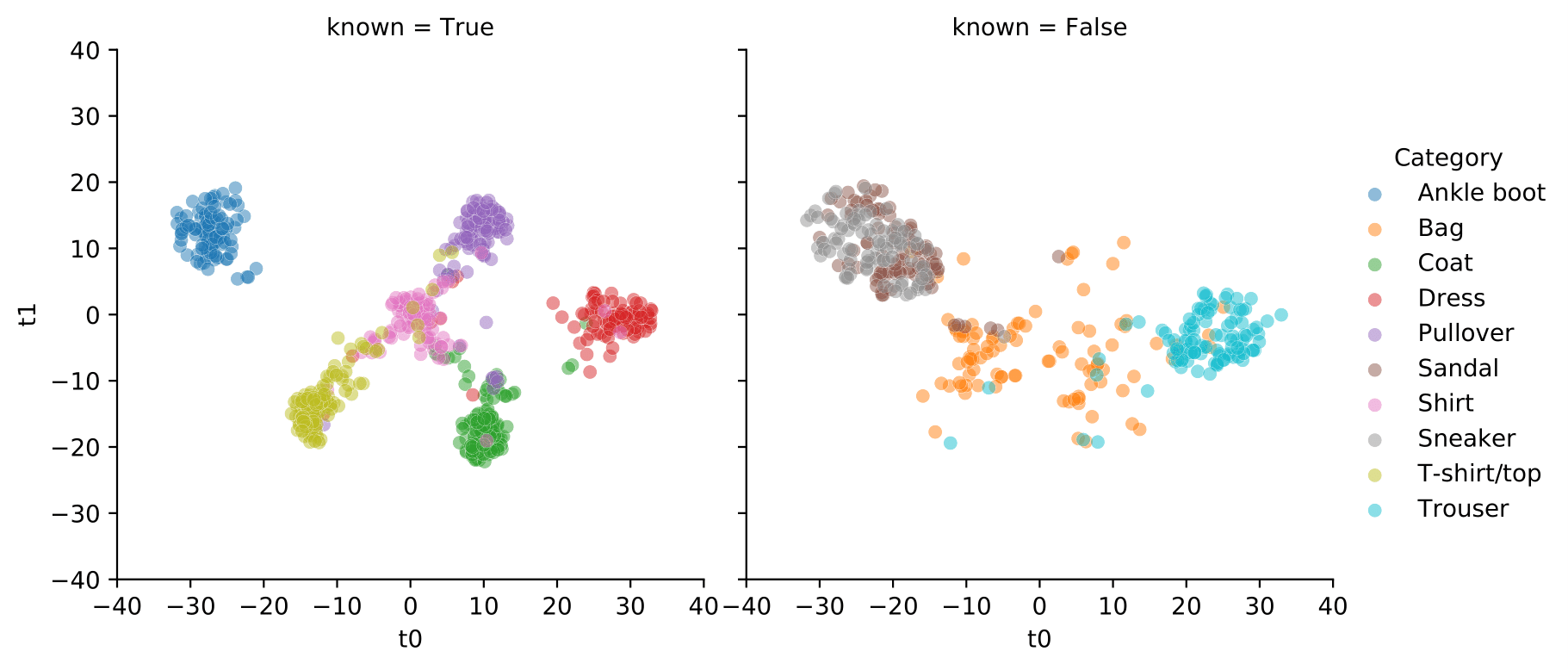}
\caption{After fine-tuning (DTAE + CE)}
\label{fig: fmnist-ft-dtae-tsne}
        \end{subfigure}%

\caption{The t-SNE plots of the Fashion-MNIST test set using cross-entropy loss: (a) training without pre-training (CE); (b)-(c): pre-training with RotNet (RotNet + CE); (d)-(e): pre-training with DTAE (DTAE + CE). The left subplots are the representations of the known class, and the right plots are the representations of the unknown classes.}
\label{fig: fmnist-ce-tsne}
\end{figure*}

To analyze the differences in representations after pre-training and
after fine-turning, we plot 1000 samples from the Fashion-MNIST test set
in Figure \ref{fig: fmnist-ce-tsne}. In these experiments, classes ``T-shirt/top'', ``Pullover'', ``Dress'', ``Coat'', ``Shirt'' and ``Ankle boot'' are known classes while the remaining classes are unknown and absent from training set. Figure \ref{fig: fmnist-wo-pt-tsne} shows the t-SNE plot of the representations learned from cross-entropy loss without pre-training. Figures \ref{fig: fmnist-pt-rotnet-tsne} and \ref{fig: fmnist-ft-rotnet-tsne} are the learned representations of the model pre-trained by RotNet and fine-tuned by cross-entropy in different stages. Figures \ref{fig: fmnist-pt-dtae-tsne} and \ref{fig: fmnist-ft-dtae-tsne} are the learned representations of the model pre-trained by DTAE and again, fine-tuned by cross-entropy. From all the final representations of the three models in Figures \ref{fig: fmnist-wo-pt-tsne}, \ref{fig: fmnist-ft-rotnet-tsne} and \ref{fig: fmnist-ft-dtae-tsne}, we observe overlaps between the known class ``Ankle boot'' (blue) and one component of the unknown class ``Sneaker'' (gray) as well as class ``Dress'' (red) and class ``Trouser'' (cyan). And the pre-training reduces the overlaps between ``Shirt'' (pink) and ``Bag'' (orange). Moreover, for the representations after pre-training, it shows that the representations learned by DTAE in Figure \ref{fig: fmnist-pt-rotnet-tsne} are more separable than those learned by RotNet in Figure \ref{fig: fmnist-pt-dtae-tsne} for different classes. Note that DTAE, similar to RotNet, is not provided with class labels, but it can find representations that are more separable among the classes than RotNet.

\begin{figure*}[t]
 \begin{subfigure}[b]{0.33\textwidth}
               \includegraphics[width=\linewidth]{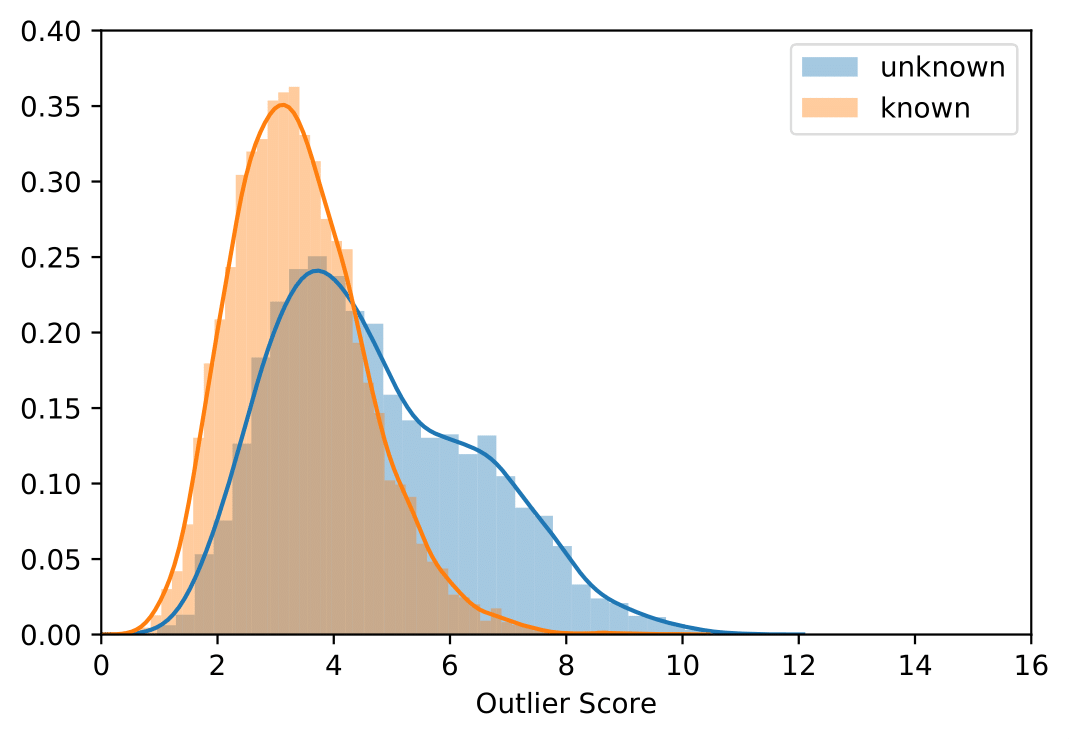}
                \caption{Without pre-training \\   (CE)}
                \label{fig: fmnist-wo-pt-score}
        \end{subfigure} 
  \begin{subfigure}[b]{0.33\textwidth}                \includegraphics[width=\linewidth]{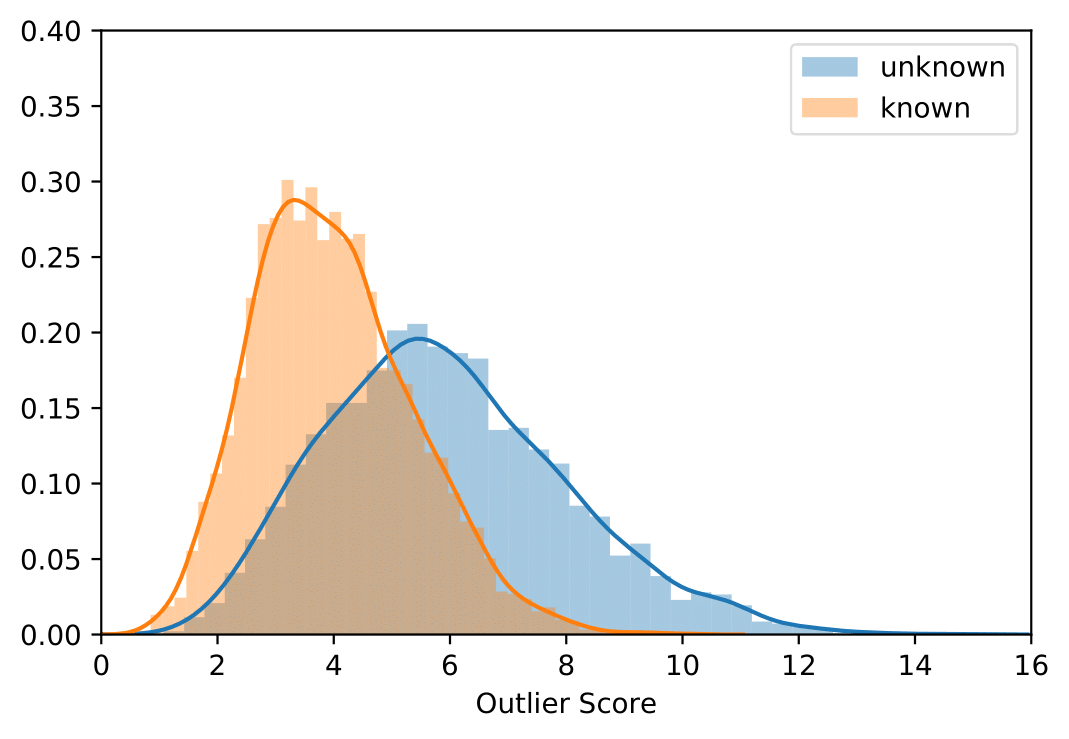}
                \caption{After fine-tuning \\ (RotNet + CE)}
                \label{fig: fmnist-rotnet-score}
        \end{subfigure}% 
  \begin{subfigure}[b]{0.33\textwidth}                \includegraphics[width=\linewidth]{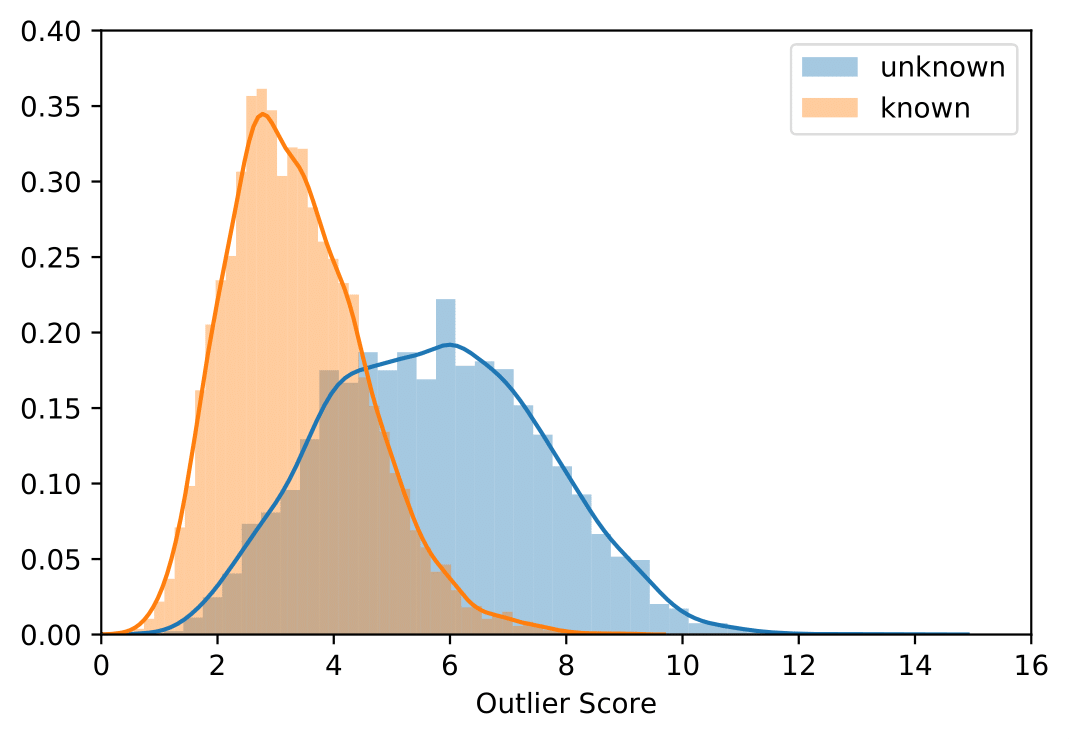}
                \caption{After fine-tuning \\ (DTAE + CE)}
                \label{fig: fmnist-dtae-score}
        \end{subfigure}% 

\caption{The distributions of outlier scores for the known and unknown classes of the Fashion-MNIST dataset in different experiments using cross-entropy loss.}
\label{fig: fmnist-ce-score}
\end{figure*}

Figure \ref{fig: fmnist-ce-score} shows distributions of the outlier scores in experiments on the Fashion-MNIST test set. Compared with the model without pre-training in Figure \ref{fig: fmnist-wo-pt-score}, the pre-training steps in Figures \ref{fig: fmnist-rotnet-score} and Figure \ref{fig: fmnist-dtae-score} increase the outlier scores in the unknown classes, which pushes their score distributions further away from the known classes. The fact that there are fewer overlaps between the known classes and the unknown class makes them more separable. The results indicate that the model pre-trained with DTAE has the fewest overlaps between the known and unknown classes.

\section{Conclusion}
\label{sec: conclude}
In this work, we introduce the self-supervision technique to OSR problems. We provide experiments across different image datasets to measure the benefits of the pre-training step for OSR problems. Moreover, we have presented a novel method: Detransformation Autoencoder (DTAE) for self-supervision. The proposed method engages in learning the representations that are invariant to the transformations of the input data. We evaluate the pre-trained model with both classification and representation loss functions. The experiments on several standard image datasets show that the proposed method significantly outperforms the baseline methods and other self-supervision techniques. Our analysis indicates that DTAE can yield representations that contain some target class information even without class labels.

\bibliographystyle{splncs04}
\bibliography{ref}

\end{document}